\documentclass[sigconf,natbib=true,anonymous=False]{acmart}
\AtBeginDocument{%
  }

\copyrightyear{2023}
\acmYear{2023}
\setcopyright{acmlicensed}\acmConference[CIKM '23]{Proceedings of the 32nd ACM International Conference on Information and Knowledge Management}{October 21--25, 2023}{Birmingham, United Kingdom}
\acmBooktitle{Proceedings of the 32nd ACM International Conference on Information and Knowledge Management (CIKM '23), October 21--25, 2023, Birmingham, United Kingdom}
\acmPrice{15.00}
\acmDOI{10.1145/3583780.3615150}
\acmISBN{979-8-4007-0124-5/23/10}

\usepackage{amsmath,multirow,booktabs,tabularx}
\usepackage{array} 
\usepackage{longtable}
\usepackage{booktabs}
\usepackage{float}
\usepackage{anyfontsize}
\usepackage{color} %颜色
\usepackage{algorithm}  
\usepackage{algorithmicx}  
\usepackage{algpseudocode}  
\usepackage{amsmath} 
\usepackage{amsfonts}
\usepackage{graphicx}
\usepackage{bm}
\usepackage{subfigure}
\usepackage{parskip}
\usepackage{multirow}
\usepackage{bigstrut}

\def\bs{\boldsymbol}
\begin{document}

\title{Bridging the KB-Text Gap: Leveraging Structured Knowledge-aware Pre-training for KBQA}
% Unlocking Semantic Perspective: Multi-Task Semantic
% Decomposition Framework for Few-Shot NER with Task-specific
% Pre-training
\author{Guanting Dong}
\authornote{Work done during internship at Meituan Inc. The first two authors contribute equally.}
\email{dongguanting@bupt.edu.cn}

\affiliation{%
  \institution{Beijing University of Posts and Telecommunication}
  \city{Beijing}
  \country{China}
}

\author{Rumei Li}
\authornotemark[1]
\email{lirumei@meituan.com}
\affiliation{%
  \institution{Meituan Group}
  \city{Beijing}
  \country{China}
}

\author{Sirui Wang}
\email{wangsirui@meituan.com}
\affiliation{%
  \institution{Meituan Group}
  \city{Beijing}
  \country{China}
}

\author{Yupeng Zhang}
\email{G0vi_qyx@buaa.edu.cn}
\affiliation{%
  \institution{Beijing University of Aeronautics and Astronautics}
  \city{Beijing}
  \country{China}
}

\author{Yunsen Xian}
\email{xianyunsen@meituan.com}
\affiliation{%
  \institution{Meituan Group}
  \city{Beijing}
  \country{China}
}

\author{Weiran Xu}
\email{xuweiran@bupt.edu.cn}
\authornote{Weiran Xu is the corresponding author}

\affiliation{%
  \institution{Beijing University of Posts and Telecommunication}
  \city{Beijing}
  \country{China}
}

% \affiliation{%

% \affiliation{%

% \affiliation{%

% \affiliation{%

% \affiliation{%

% \affiliation{%

% \affiliation{%

\renewcommand{\shortauthors}{Guanting Dong et al.}

\begin{abstract}
Knowledge Base Question Answering (KBQA) aims to answer natural language questions with factual information such as entities and relations in KBs. However, traditional Pre-trained Language Models (PLMs) are directly pre-trained on large-scale natural language corpus, which  poses challenges for them in understanding and representing complex subgraphs in structured KBs. To bridge the gap between texts and structured KBs, we propose a \textbf{S}tructured \textbf{K}nowledge-aware \textbf{P}re-training method (SKP). In the pre-training stage, we introduce two novel structured knowledge-aware tasks, guiding the model to effectively learn the implicit relationship and better representations of complex subgraphs. 
In downstream KBQA task, we further design an efficient linearization strategy and an interval attention mechanism, which assist the model to better encode complex subgraphs and shield the interference of irrelevant subgraphs during reasoning respectively.
Detailed experiments and analyses on WebQSP verify the effectiveness of SKP, especially the significant improvement in subgraph retrieval (+4.08\% H@10).
\end{abstract}


% \ccsdesc[500]{Understanding multi-modal content~natural language processing}
% \ccsdesc[100]{Information access and retrieval~retrieval models}

% \ccsdesc[300]{Integration and aggregation~knowledge graphs}

\maketitle

\section{Introduction}

Knowledge Base Question Answering (KBQA)  aims to seek answers to factoid questions from structured KBs \cite{auer2007dbpedia,bollacker2008freebase}. The existing methods to solve KBQA can be divided into two categories: Semantic parsing-based (SP-based) methods \cite{ye2021rng,das2022knowledge,lan2020query,sun2020sparqa} and embedding-based methods \cite{agarwal-etal-2021-knowledge,zhang2022subgraph,he2021improving,sun2020faithful,sun2018open,sun2019pullnet}. The former one heavily relies on the expensive annotation of the intermediate logic form such as SPARQL \cite{perez2009semantics}. Instead of parsing the questions, the latter one directly encodes and retrieves the candidate subgraphs, then obtains answers by a ranker or a generator \cite{saxena2022sequence,oguz2020unik,yu2022decaf}. Therefore, embedding-based framework is more suitable for the realistic dialogue system, which has gained attention in the research field.

%intro引出问题
Unfortunately, the existing embedding-based methods still encoun-\\
ter challenges in understanding and reasoning with structured knowledge. Traditional KBs, such as Freebase, typically comprise sets of subject-predicate-object (SPO) triplets. However, the PLMs are pre-trained on unstructured natural language texts, making it difficult to encode the implicit logical relationship in the structured subgraphs \cite{yang2020improving}. Moreover, to improve the efficiency, existing methods directly concatenate all the retrieved subgraphs during answer generation, which inevitably causes different types of subgraphs to interfere with each other in the encoding process \cite{de2022fido,yu-etal-2022-kg}.

% However, the existing embedding-based methods  still suffer from the problem of understanding structured knowledge, which can be divided into the following two parts:
% (1) \textbf{Understanding of subgraphs.} The subgraphs in the large-scale structured KBs (e.g. Freebase) are usually represented as a set of subject-predicate-object (SPO) triplets. However, traditional PLMs are pre-trained on diverse unstructured natural language texts, which is difficult for model to understand the implicit relationship in the subgraph\cite{yang2020improving}.
% (2) \textbf{Encoding subgraphs for reasoning.} As the retriever-reader architecture \cite{saxena2022sequence,oguz2020unik,yu2022decaf} gradually become the mainstream framework of KBQA, how to improve encoding subgraphs for reasoning has also become a big challenge. During retrieving subgraphs, a large amount of complex subgraphs in KBs contain n-ary relations, which are difficult for retriever to completely represent the logical relationship of these complex subgraphs. 
% Moreover, traditional methods directly concatenate the all the retrieved subgraphs during generating answers, which inevitably cause different types of subgraphs to interfere with each other in the encoding process.
% diss previous work
%--------------main 图-----------------
\begin{figure*}
    \centering
    \resizebox{0.90\textwidth}{!}{
    \includegraphics{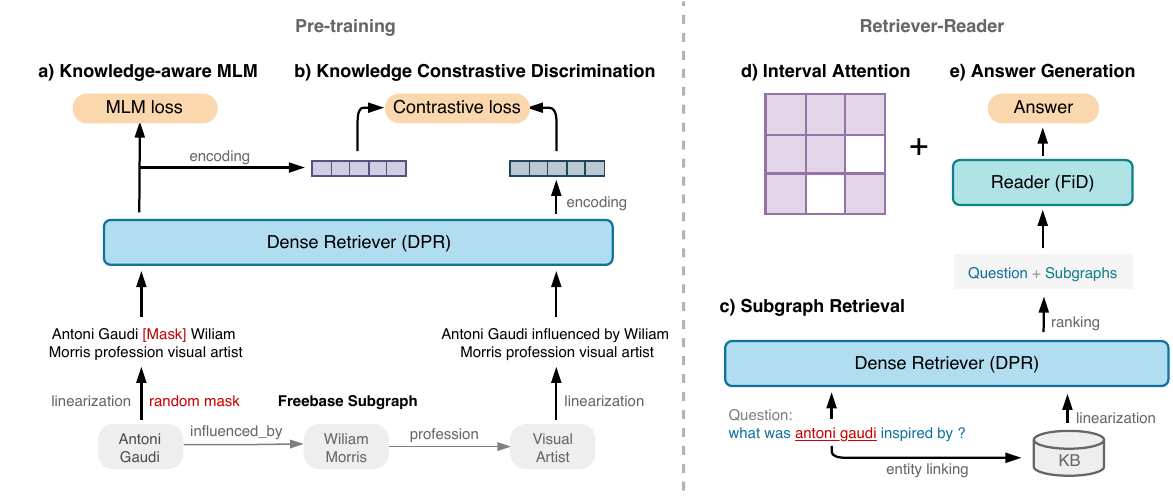}}
    \vspace{-0.3cm}
    \caption{The overall architecture of our proposed SKP}
    \label{fig:main}
    \vspace{-0.2cm}
\end{figure*}
%--------------main 图-----------------
Previous works \cite{bordes2013translating,ji2015knowledge,sun2018open,sun2019pullnet,zhao-etal-2022-entity,zhang-etal-2023-pay,ju2022grape} primarily addresses these challenges by introducing external text corpus and specially designed framework to incorporate information from the documents. However, the required external resources may be hard to collect in practice. 
Besides, another branch of case-based methods \cite{das2022knowledge,ye2021rng,das2021case,yu-etal-2022-kg} integrate training cases or candidate subgraphs to prompt the model for reasoning. Nevertheless, these methods easily trigger additional noise and are limited by the data quality.
Therefore, it is essential to develop a method for PLMs to improve its ability of understanding structured knowledge and anti-interference\cite{dong2022pssat}, which facilitates techniques of KBQA to be widely applied to the realistic question answering and task-oriented dialog system \cite{zeng2022semisupervised,qixiang-etal-2022-exploiting,yuan2023scaling,10193387}.

To address the above limitations, we propose \textbf{S}tructured \textbf{K}nowledge-aware \textbf{P}re-training method (SKP) to enhance the model's ability of understanding structured knowledge and encoding complex subgraphs. Inspired by masked language modeling (MLM) \cite{devlin2018bert} and contrastive learning \cite{chen2020simple,9747192,10094766,10095149}, we introduce two novel structured knowledge-aware pre-training tasks according to the characteristics of the data in subgraphs. These tasks effectively facilitate the learning of implicit relationships and representations in structured knowledge.
In the downstream KBQA task, we follow the previous retriever-reader framework, including subgraphs retrieval and answer generation. What distinguishes our approach is the introduction of an effective linearization strategy that significantly reduces the number of candidate subgraphs in the KB while preserving the structural semantic information. Furthermore, we initialize the retriever with pre-trained parameters to facilitate the transfer of upstream learned structured knowledge to the downstream model.  To better encode different subgraphs during reasoning, motivated by \citeauthor{liu2020k}, we design an interval attention mechanism, effectively guiding the model to shield the interference from irrelevant subgraphs. Our contributions are as follows: 

(1) We propose two novel structured knowledge-aware pre-training tasks to enhance the learning of implicit relationships and representations in subgraphs for KBQA. (2) We design an efficient linearization strategy and an interval attention mechanism to improve the model's ability to encode complex subgraphs and mitigate interference from irrelevant subgraphs during reasoning. (3) Experiments and analyses on WebQSP show the effectiveness of SKP, especially huge improvements in the subgraph retrieval. Our source codes and datasets are available at Github\footnote{https://github.com/dongguanting/SKP-for-KBQA, We will further release the results and code of SKP with LLMs for comparison.}.

\section{Method}
\subsection{Problem Definition}
In this paper, we mainly focus on the embedding-based KBQA. Given a query $q$, we first annotate the named entities in the query and link them to the nodes in the KB. Then some heuristic algorithm is applied to retrieve a set of query specified subgraphs $G = \{(e, r, e^{'}) |e, e^{'} \in E, r \in R\}$, where $E$ and $R$ denote the entity set and the relation set respectively. Our task is to figure out the answers $A_q$ to the query $q$ from retrieved subgraphs $G$. 

% In this section, we first present two structured knowledge pre-training tasks, then we introduce the overall framework of our proposed SKP.
\subsection{Structured Knowledge-aware Pre-training}
The performance of KBQA depends heavily on understanding the implicit relationship of subgraphs . To further enhance the relation learning of PLMs, we propose two novel pre-training tasks, which are shown in Figure \ref{fig:main}(left).

\textbf{Knowledge-aware MLM (KM).} We follow the design of masked language modeling (MLM) in BERT \cite{devlin2018bert}. Given a subgraph $g \in G$, we linearize $g$ in the form of natural languages, which can be formulate as "$\mathtt{[CLS]} \,  e\, r\, e' \, \mathtt{[SEP]}$".
%对复杂子图加角注解释
For each subgraph $g$, we randomly replace one of the subject $e$, relation $r$ or object $e^{'}$ with the special [MASK] symbol, and then try to recover them. It's worth noting that if the masked entity consists of multiple tokens, all of the component tokens will be masked.
The purpose of our design is to guide the PLMs to learn the implicit relationship between triplets. Hence, the loss function of the MLM is:
\begin{equation}
\begin{small}
{L}_{mlm}=-{\sum}_{m=1}^{M}\log P({x_m})
\end{small}
\end{equation}
where $M$ is the total number of masked tokens and $P(\bm{x_m})$ is the predicted probability of the token $x_m$ over the vocabulary size.

\textbf{Knowledge Contrastive Discrimination (KCD).} To better discriminate different subgraph representations in semantic space, we introduce Knowledge Contrastive Discrimination. Specifically, given a subgraph $g$, for postive samples, we randomly replace one of the subject $e$, relation $r$ or object $e^{'}$ with the special [MASK] symbol. For negative samples, we adopt other samples in the batch. We use $h_{o}$, $h_{p}$, $h_{n}$ to indicate the representations of the original, the positive, and the negative input samples respectively. Meanwhile, we employ InfoNCE \cite{aitchison2021infonce} objective by bringing original samples $h_{o}$ closer to their semantically similar positive samples $h_{p}$ and further away from negative samples $h_{n}$. We formulate $L_{C}$ as follows:
\begin{equation}
\small
\mathcal L_{\rm C}=-\frac{1}{N}\sum_{i=1}^{N} \log\frac{\exp\{S(\bs h_{\bs oi}, \bs h_{\bs pi})/\tau\}}{\sum_{j=1}^{N-1}\exp\{S(\bs h_{\bs oi}, \bs h_{\bs nj})/\tau\}} 
\end{equation}
where $N$ denotes the number of total examples in the batch. $\tau$ is a temperature hyperparameter and $S(\cdot)$ is cosine similarity. Hence, we obtain the overall loss function L of our joint pre-training tasks:
\begin{equation}
% \begin{aligned}
\begin{small}
L = \alpha L_{mlm}+ (1-\alpha) L_{C}
\end{small}
\end{equation}
where $L_{mlm}$ and $L_C$ denote the loss functions of the two tasks. In our experiments, we set $\alpha = 0.6$.
\subsection{Downstream KBQA task}
% We use a retriever-reader architecture, with dense passage retriever (DPR)\cite{karpukhin2020dense} as retriever and fusion-in-decoder (FiD)\cite{izacard2020leveraging} as our reader. We first propose an efficient linearization strategy for complex subgraphs in KB to convert structured knowledge to text. Then, we use DPR to retrieve from heterogeneous text passages and concatenate each retrieved document with the question. Finally, we integrate our new attention mechanism into the reader encoder and generate the answer according to the retrieved document. The overall architecture is illustrated in Figure x.
We adopt a retriever-reader architecture, with dense passage retriever (DPR) 
 \cite{karpukhin2020dense} as retriever and fusion-in-decoder (FiD) \cite{izacard2020leveraging} as our reader. The overall architecture is illustrated in Figure \ref{fig:main}.
%--------------------------linear图-----------------------
\begin{figure}[t]
\centering

\resizebox{.36\textwidth}{!}{\includegraphics{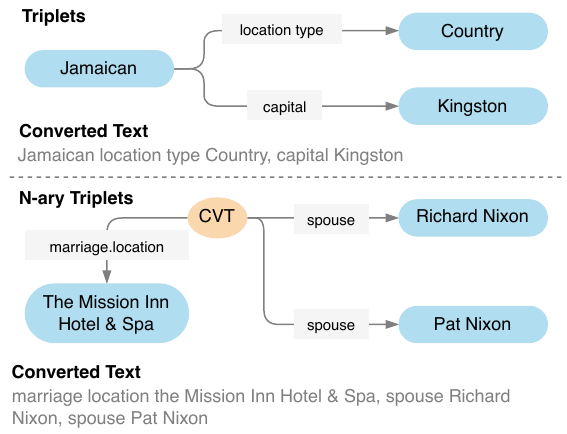}}
\caption{Knowledge base linearization. We show examples of how we linearize valina triplets and complex triplets (multiple entities and relations with a central CVT node).}
\vspace{-0.2cm} 
\label{fig:linear}
\vspace{-0.2cm} 
\end{figure}
%--------------------------linear图-----------------------

\textbf{Linearization.} Traditional KBs, which contains triplets composed of subject, predicate and object, are usually stored in Resource Description Format (RDF). For the vanilla triples which occupies 90\% of the total number of triples in KB, previous works \cite{das2022knowledge,yan2021large,yu2022decaf} simply flattens each triple without any merging and pruning. Different from them, our linearization strategy can be divided into 6 cases: (1) same subject and predicate, (2) same subject and object (3) same predicate and object (4) same subject (5) same predicate, and (6) same object. To preserve more information when converting all vanilla triples into subgraphs, we sequentially traverse the entire KB and prune triples based on the priority order of cases 1 to 6. We use 100 tokens as the threshold for converting and ensure that hard truncation does not occur. Each subgraph typically contains 2-3 vanilla triples . For complex(n-ary) triplets, freebase uses compound value types (CVTs) to convert a n-ary relation into multiple standard triplets. To better represent them, we convert a n-ary relation into a single sentence by forming a comma-separated clause for each predicate (Figure \ref{fig:linear}). In this way, linearized subgraphs preserve structural semantic information while greatly reduce the number of candidate subgraphs in KB (from 230 million to 112 million).

% In addition, KBs contain complex(n-ary) triples that express complicated relationship. The linearization details for two types of triples are discussed in Appendix \ref{app:linear} and Figure \ref{fig:linear}. In this way, linearized subgraphs preserve structural semantic information while greatly reduce the number of candidate subgraphs in KB (from 230 million to 112 million).

 % 2). For complex(n-ary) triplets, freebase uses compound value types (CVTs) to convert an n-ary relation into multiple standard triplets. To better represent them, we convert an n-ary relation into a single sentence by forming a comma-separated clause for each predicate. In this way, linearized subgraphs preserves structural semantic information, while greatly reducing the number of candidate sub
 % graphs in KB (from 230 million to 112million ).

\textbf{Retriever.} Dense Passage Retriever (DPR) applies two fine-tuned BERTs \cite{devlin2018bert} to encode passages and questions respectively, and then retrieves the top-k passages based on the dot-product similarity. It further utilizes FAISS \cite{johnson2019billion} to conduct an similarity search. Through this step, we can retrieve $N$ passages which are potentially relevant to the input question. We follow the standard setup of DPR for fine-tuning. One difference is that we initialize the retriever with pre-trained parameters, which transfer the upstream learned structured knowledge into downstream retriever. Detailed hyperparameter settings can be found in the implementations.

\textbf{Reader.} The Fusion-in-Decoder\cite{izacard2020leveraging} reader has demonstrated to be strongly effective in fusing information from a large number of documents. However, previous works directly perform cross-attention over concatenation of retrieved passages and questions, which inevitably causes retrieved subgraphs to interfere with each other in encoding process \cite{de2022fido}. To alleviate the above problem, we introduce \textbf{Interval Attention Mechanism (IAM)}. As shown in Figure \ref{fig:main}(d), through setting 1 or 0 for the cross-attention matrix, we make input question and all retrieved subgraphs visible to each other, and different types of subgraphs invisible to each other. We utilize the FiD with T5-large \cite{raffel2020exploring} and further integrate IAM into FiD encoder. We follow the original setting for all experiments which are introduced in implementation details.

\begin{table}[!tbp]
\centering
  \resizebox{0.45\linewidth}{!}{
  \begin{tabular}{l|c}
    \toprule
    \multirow{1}[2]{*}{\textbf{Methods}} & \multicolumn{1}{c}{\textbf{Hits@1}} \\
    \midrule
    GraftNet \cite{sun2018open} & 69.5 \\    
    PullNet \cite{sun2019pullnet} & 68.1 \\
    EMQL \cite{sun2020faithful}& 75.5\\
    Bert-KBQA \cite{yan2021large} & 72.9 \\
    NSM \cite{he2021improving} & 74.3 \\
    KGT5 \cite{saxena2022sequence} & 56.1 \\
    SR-NSM \cite{zhang2022subgraph} & 69.5 \\
    CBR-SUBG \cite{das2022knowledge} &72.1 \\
    EmbededKGQA \cite{saxena2020improving}& 72.5 \\
    DECAF(Answer only) \cite{yu2022decaf} &  74.7 \\
    UniK-QA$^{*}$ \cite{oguz2020unik} &  77.9 \\
    \midrule
   \textbf{SKP} (Ours)& \textbf{79.6} \\
    \hspace{0.5cm} w/o IAM & 79.3 \\
    \hspace{0.5cm} w/o KCD & 79.1 \\
    \hspace{0.5cm} w/o KM & 78.7 \\
   \hspace{0.5cm} w/o KM+KCD & 78.4 \\
   \hspace{0.5cm} w/o KM+KCD+IAM & 77.9 \\
    \bottomrule
    \end{tabular}
    }
    % \vspace{-0.2cm} 
    \caption{Experimental results on WebQSP dataset. “w/o" denotes the model performance without a specific module. * denotes the reproduced result.}
    
  \label{tab:main}%
  \vspace{-0.6cm}
\end{table}%
 % Baseline results are taken from \cite{yan2021large,yu2022decaf}
%-----------main result -----------

\section{Experiments}
\textbf{Datasets and Metric.} To evaluate our SKP, we conduct several experiments on WebQuestionsSP \cite{yih2016value} dataset which contains 4737 questions that are answerable using Freebase. We follow \citeauthor{sun2018open} and utilize 2848/250/1639 examples for training, validation, and test respectively. 
For QA performance, we use Hits@1 as our evaluation metric, the percentage of the model’s top-predicted answer being a “hit” (exact match) against one of the gold-standard answers.

\textbf{Implementation Details.} For the structured pre-training stage, we use BERT-base-uncased \cite{devlin2018bert} as backbone. We set the batch size of BERT to 8 and the pre-training takes an average of 12 hours for 5 epochs. The corresponding learning rates are set to 1e-5. For the pre-training corpus, We randomly selected 1 million linearized subgraphs from Freebase as training data, which means no external KBs will be introduced in the pre-training stage. In order to align with previous work, we use STAGG \cite{yih2015semantic} as entity linking system to narrow down the search to a high-recall 2-hop neighborhood of the retrieved entities for each question. Then we convert KB relations in the 2-hop neighborhood into text, retrieve the most relevant ones using DPR to form 100 context passages, and feed them into the FiD. We also align the settings with UniK-QA \cite{oguz2020unik}.

\textbf{Baselines.}
We mainly compare SKP with multiple state-of-the-art baselines
that use embedding-based method for KBQA as follow:
\textbf{GraftNet} \cite{sun2018open} 
, \textbf{PullNet} \cite{sun2019pullnet}
, \textbf{Bert-KBQA} \cite{yan2021large} 
, \textbf{EmbedKGQA} \cite{saxena2020improving}
, \textbf{NSM} \cite{he2021improving} 
, \textbf{SR-NSM} \cite{zhang2022subgraph}
, \textbf{EmQL} \cite{sun2020faithful} 
, \textbf{KGT5} \cite{saxena2022sequence} 
, \textbf{CBR-SUBG} \cite{das2022knowledge} 
, \textbf{UniK-QA} \cite{das2022knowledge} 
, \textbf{DeCAF} \cite{yu2022decaf}
, \textbf{DPR} \cite{karpukhin2020dense}. Specifically, for DPR, there are four variants as follows: (1) \textbf{DPR(NQ)}: DPR trained on training set of NQ dataset. (2) \textbf{DPR(NQ-adv)}: DPR trained on training set of NQ dataset and adversarial hard negatives samples. (3) \textbf{DPR(Multi)}: DPR trained on training set of Multi-dataset.(4) \textbf{DPR(WebQ)}: DPR trained on training set of WebQ dataset. It is worth noting that \citeauthor{oguz2020unik} has opened all the DPR variants.
% % \textbf{Retriever.} DPR applies the passage encoder and the question encoder to encode the all passages and questions, and retrieve the top-k passages based on the dot-product similarity. Then it applies FAISS\cite{johnson2019billion} to conduct an efficient similarity search retrieve passages which are potentially relevant to the input question.
% \textbf{Retriever.} We initialize the DPR with pre-trained parameters and train DPR using the standard setup \cite{karpukhin2020dense} for retrieval. We train for 40 epochs with a linear warmup of 500 steps, batch size of 128 and learning rate 10e-5.

% \textbf{Reader.} We adopt the FiD model with T5-large \cite{raffel2020exploring} and 100 context documents and use the original hyper-parameters of FiD \cite{izacard2020leveraging} whenever possible. In particular, the Adam \cite{kingma2014adam} optimizer is used with a constant learning rate of 0.0001. The model is trained for 10k steps, with a batch size of 64, using 16 A100 GPUs. We did not perform any hyper-parameter search. We will release our code after blind review. 

% \footnote{Due to space limitations, detailed introductions of implementation, baselines are shown in Appendix }.

% %-----------speciail study-----------
% \begin{figure}[t]
% \centering
% \resizebox{.48\textwidth}{!}{\includegraphics{figures/50KB_3.png}
% % \vspace{-0.2cm} 
% \caption{The performance comparison with the full KB and the 50\% KB. We only compare to baselines that also report results with 50\% KB.}
% \label{fig:6}
% \vspace{-0.4cm} 
% \end{figure}

% %-----------speciail study-----------

\subsection{Main Result}
Table \ref{tab:main} shows the main results compared to different embedding-based baselines on the WebQSP.  Results show that our proposed SKP greatly outperforms all the baselines on Hit@1,  especially 1.7\% higher than the SOTA method (UniK-QA). Besides, SKP has a significant performance degradation when removing both KM and KCD (79.5 to 78.4), showing that structured knowledge-aware pre-training tasks benefit the gap between natural language texts and structured subgraphs.

%------------------------retrieval--------------------------------
\begin{table}[t]
  \centering
  \resizebox{0.9\columnwidth}{!}{
  \begin{tabular}{lcccc}
    \toprule
    \multirow{2}{*}{\textbf{Methods}}&\multicolumn{3}{c}{WebQSP} \\
    \cmidrule(lr){2-5}
    &Hits@10& Hits@20 & Hits@50 & Hits@100 \\
    \midrule
    \multicolumn{1}{c}{\textit{Baselines}} \\
    Bert-KBQA&35.91&48.74&66.9&77.03 \\
    DPR(NQ) &80.96& 86.21 &91.52 & 93.11 \\
    DPR(NQ-adv)&80.93& 80.64 &91.76 & 93.10 \\
    DPR(Multi)&87.22 &91.64 &93.95 &94.68 \\
    DPR(WebQ)&87.46 &91.92 &94.11 &94.87 \\
    UniK-QA &87.50 &- &- &95.85 \\
    \midrule
    \multicolumn{1}{c}{\textit{Our implementations}} \\
    SKP(NQ)&91.23 &93.35 &95.64 &96.32 \\
    SKP(WebQ) &\textbf{91.58} &\textbf{93.65} &\textbf{95.85} &\textbf{96.64} \\
    \bottomrule
%-------------------------retrieval-------------------------------------

   \vspace{-0.2cm} 
  \end{tabular}}
  \caption{Subgraph retrieval results on WebQSP. The content in parentheses denotes retriever fine-tuned by specific dataset . We get the retrieval result of UniK-QA from \citeauthor{oguz2020unik}.}
  \label{tab:retrieve}
  \vspace{-0.5cm}
  
\end{table}
% --------------------------retrieval----------------------------
\textbf{Ablation Studies.} We conduct an ablation study in table \ref{tab:main} to investigate the characteristics of the main components in SKP.  We can observe that: 1) the performance of SKP drops when removing any component, which suggests that every part of design is necessary. 2) The joint training of two pre-training tasks can promote each other, and even training them individually (79.2\% and 78.9\%) can outperform all the baselines by a certain margin. 3) The SKP only with interval attention mechanism has a positive effect as well (from 77.9 to 78.4), which proves the necessity of removing interference of subgraphs in the generation stage.

\textbf{Retrieval experiment.} To validate the effectiveness of SKP on understanding structured knowledge, we evaluate the subgraph retrieval results on WebQSP. We initialize our retriever with pre-trained parameters and fine-tune on the training samples of specific dateset. As shown in Table \ref{tab:retrieve}, our method far exceeds all the baseline methods in all settings (+4.08\% H@10). To align with the UniK-QA, we also fine-tune our method on the NQ dataset, which still outperforms the best baseline by 3.73\%(H@10) and 0.47\%(H@100). The results fully verify the effectiveness of our approach.
% It is worth noting that our linearization strategy prunes and merges half the size of KB(230 million->112million ), while still maintaining an excellent retrieval score, which fully proves the effectiveness of our approach.

\begin{figure}[t]
\centering
\resizebox{.45\textwidth}{!}{\includegraphics{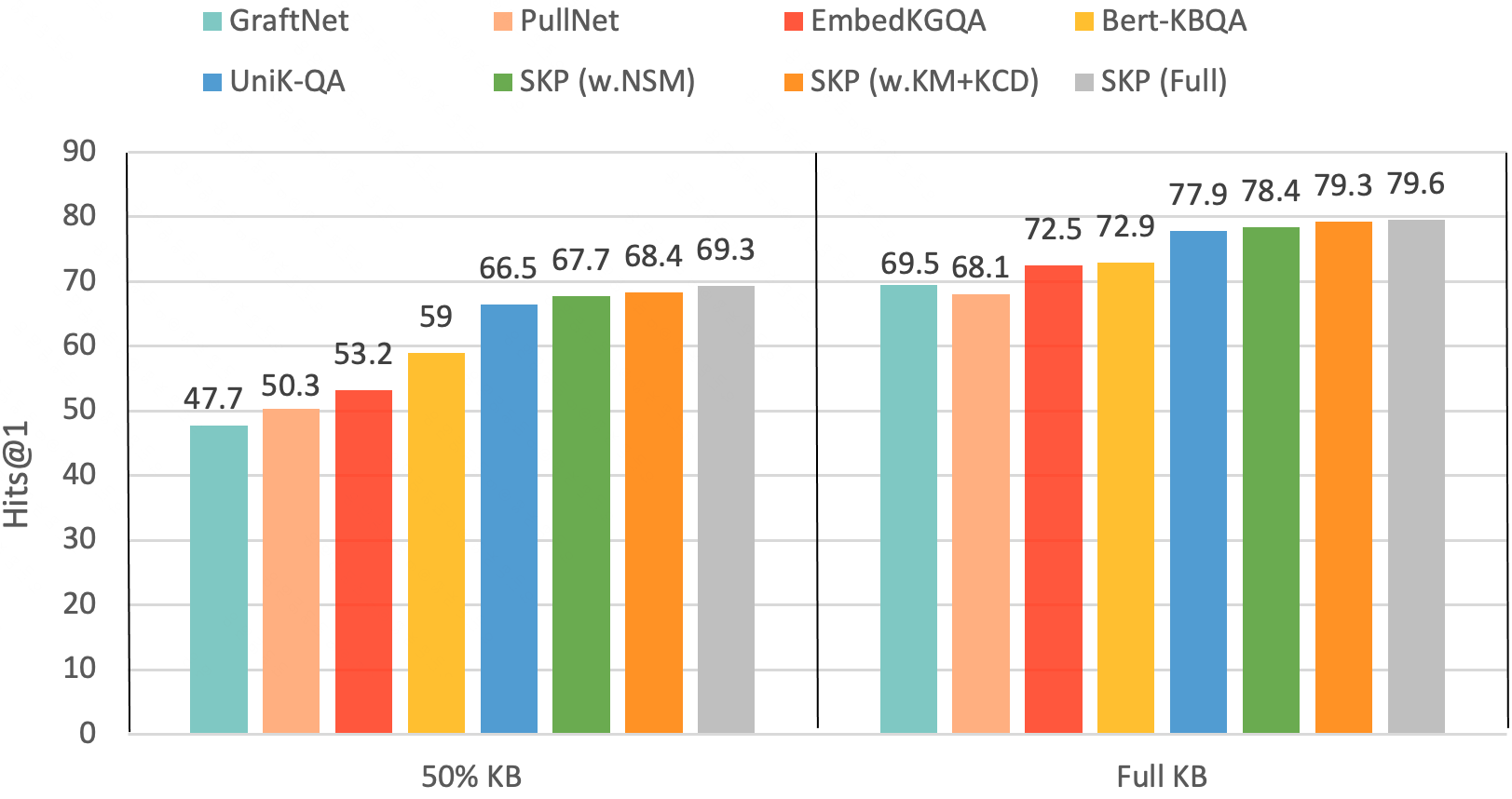}}
\vspace{-0.3cm} 
\caption{The performance comparison with the full KB and the 50\% KB. "w." denotes the SKP with specific component.}
\label{fig:50kb}
\vspace{-0.1cm} 
\end{figure}

\textbf{Results over the Incomplete KB.} To further explore the robustness of our approach over the incomplete KB, we randomly remove 50\% of the KB facts in the retrieved subgraphs and conduct experiments on this incomplete version of the WebQSP dataset. The results are illustrated in Figure \ref{fig:50kb}. We can observe that: 1) Our approach consistently outperforms other baselines under both the full KB and the 50\% KB settings. 
2) With 50\% KB, adding structured knowledge-aware pre-training tasks and interval attention mechanism achieve more performance gain than with full KB (+1.6 vs +1.1/ + 0.9 vs +0.2), demonstrating the generalization ability of SKP in low-resource scenarios.
%-----------speciail study-----------

%-----------speciail study-----------

% %--------------------------attention图-----------------------

% \begin{figure}[t]
%     \centering
%     \subfigure[Baseline]{
%         \includegraphics[width=.22\textwidth]{figures/output_unmask.pdf}
%         \label{label_for_cross_ref_1}
%     }
%     \subfigure[SKP]{
% 	\includegraphics[width=.22\textwidth]{figures/output_mask.pdf}
%         \label{label_for_cross_ref_2}
%     }
%     \vspace{-0.4cm}
%     \caption{Visualization of the attention mechanism for FiD. }
%     \label{fig:visual}
%     \vspace{-0.2cm}
% \end{figure}

% %------------------------retrieval--------------------------------
% \textbf{Visualization.} To gain an insight into the effectiveness of the interval attention mechanism, we visualize the cross attention weights of FiD encoder, as shown in Figure \ref{fig:visual}. Specifically, we visualize the cross attention matrix of the baseline method (Unik-QA) and SKP in 10 * 10 matrix dimension. Compared with baseline method, it is obvious that high cross attention weights are basically distributed on the diagonal of the matrix, which means that our individual subgraph is more inclined to interact with itself. This finding confirms that our framework can effectively shield the interference of irrelevant subgraphs during reasoning.

\section{Conclusion}
In this paper, we propose a \textbf{S}tructured \textbf{K}nowledge-aware \textbf{P}re-training method (SKP). Specifically, we present two novel pre-training tasks, for effectively learning the implicit relationship and better representation of complex subgraphs. In downstream KBQA task, we further design an efficient linearization strategy and an interval attention mechanism, which assist the model to better encode complex subgraphs and shield the interference of irrelevant subgraphs during reasoning. Detailed experiments and analyses on WebQSP verify the effectiveness of SKP, especially the significant improvements in subgraph retrieval (+4.08\% H@10).

% \section*{Acknowledgements}
% This document has been adapted by Jordan Boyd-Graber, Naoaki Okazaki, Anna Rogers from the style files used for earlier ACL, EMNLP and NAACL proceedings, including those for
% EACL 2023 by Isabelle Augenstein and Andreas Vlachos,
% EMNLP 2022 by Yue Zhang, Ryan Cotterell and Lea Frermann,
% ACL 2020 by Steven Bethard, Ryan Cotterell and Rui Yan,
% ACL 2019 by Douwe Kiela and Ivan Vuli\'{c},
% NAACL 2019 by Stephanie Lukin and Alla Roskovskaya, 
% ACL 2018 by Shay Cohen, Kevin Gimpel, and Wei Lu, 
% NAACL 2018 by Margaret Mitchell and Stephanie Lukin,
% Bib\TeX{} suggestions for (NA)ACL 2017/2018 from Jason Eisner,
% ACL 2017 by Dan Gildea and Min-Yen Kan, NAACL 2017 by Margaret Mitchell, 
% ACL 2012 by Maggie Li and Michael White, 
% ACL 2010 by Jing-Shin Chang and Philipp Koehn, 
% ACL 2008 by Johanna D. Moore, Simone Teufel, James Allan, and Sadaoki Furui, 
% ACL 2005 by Hwee Tou Ng and Kemal Oflazer, 
% ACL 2002 by Eugene Charniak and Dekang Lin, 
% and earlier ACL and EACL formats written by several people, including
% John Chen, Henry S. Thompson and Donald Walker.
% Additional elements were taken from the formatting instructions of the \emph{International Joint Conference on Artificial Intelligence} and the \emph{Conference on Computer Vision and Pattern Recognition}.


\bibliographystyle{ACM-Reference-Format}
\bibliography{acmart}

% \appendix








% Nam interdum magna at lectus dfignissim, ac dignissim lorem

\end{document}